\documentclass[10pt,twocolumn,letterpaper]{article}

\usepackage{iccv}
\usepackage{times}
\usepackage{epsfig}
\usepackage{graphicx}
\usepackage{amsmath}
\usepackage{amssymb}
\usepackage{booktabs}
\usepackage{multirow}

\usepackage{enumitem}
\usepackage{kotex}
\usepackage[hang]{subfigure}
\usepackage[ruled]{algorithm2e}
\usepackage[accsupp]{axessibility}  
\usepackage[pagebackref=true,breaklinks=true,letterpaper=true,colorlinks,bookmarks=false]{hyperref}

\newcommand{\eqdef}{\overset{\mathrm{def}}{=\joinrel=}}

\iccvfinalcopy 

\def\iccvPaperID{10181} 

\ificcvfinal\fi

\begin{document}

\title{Influence-Balanced Loss for Imbalanced Visual Classification}

\author{Seulki Park \quad Jongin Lim \quad Younghan Jeon \quad Jin Young Choi\\
ASRI, Dept. of Electrical and Computer Engineering, Seoul National University\\
{\tt\small \{seulki.park, ljin0429, yh1992, jychoi\}@snu.ac.kr}
\and
}

\maketitle
\ificcvfinal\thispagestyle{empty}\fi

\begin{abstract}

In this paper, we propose a balancing training method to address problems in imbalanced data learning.
To this end, 
we derive a new loss used in the balancing training phase that alleviates the influence of samples that cause an overfitted decision boundary.
The proposed loss efficiently improves the performance of any type of imbalance learning methods.
In experiments on multiple benchmark data sets, we demonstrate the validity of our method and reveal that the proposed loss outperforms the state-of-the-art cost-sensitive loss methods.
Furthermore, since our loss is not restricted to a specific task, model, or training method, it can be easily used in combination with other recent re-sampling, meta-learning, and cost-sensitive learning methods for class-imbalance problems. 
Our code is made available at \url{https://github.com/pseulki/IB-Loss}.

\end{abstract}
\vspace{-3mm}
\section{Introduction}
Despite the remarkable success of deep neural networks (DNNs) these days, many areas of computer vision suffer from highly imbalanced datasets. 
Many real-world data exhibit skewed distributions \cite{ref:data_coco, ref:data_iNat, ref:data_pascal_voc, data:liu_yu_cvpr2019, data:uci}, in which the number of samples per class differs greatly. 
This imbalance between classes can be problematic, since the model trained on such imbalanced data tends to overfit the dominant (majority) classes~\cite{ref:japkowicz_stehphen_2002, ref:He_Garcia_2009, ref:buda_mazurowski_2018}.
That is, while the overall performance appears to be satisfactory, the model performs poorly on minority classes.
To overcome the class imbalance problem, extensive research has recently been conducted to improve the generalization performance by reducing the overwhelming influence of the dominant class on the model.

The research on imbalanced learning can be divided into three approaches: data-level approach, cost-sensitive re-weighting approach, and meta-learning approach. 
The data-level approach aims to directly balance the training data distributions via re-sampling (i.e., under-sampling or over-sampling) \cite{ref:resample_chawla_smote02, ref:resample_hulse_icml07} or by generating synthetic samples \cite{ref:sankha_gamo_iccv2019}. 
Meanwhile, the cost-sensitive re-weighting approach aims to design new loss functions to re-weight samples by considering their importance \cite{ref:wang_hebert_nips17, ref:huang_tang_cvpr16, ref:lin_focal_loss_iccv17}.
Finally, the meta-learning approach enhances the performance of the data-level and/or cost-sensitive re-weighting approach via meta-learning  \cite{ref:shu_metaweightnet_neurips2019, ref:liu_mesa_neurips20, ref:ren_bms_neurips2020}. 
Most recent data-level approaches require a heavy computational burden.
Moreover, under-sampling can lose some valuable information, and over-sampling or data generation can cause overfitting on certain repetitive samples. The meta-learning approach requires additional unbiased data~\cite{ref:shu_metaweightnet_neurips2019} or a meta-sampler  \cite{ref:ren_bms_neurips2020}, which is computationally expensive in practice.
Therefore, our work focuses on the cost-sensitive re-weighting approach to design a new loss function that is simple but efficient.

The cost-sensitive re-weighting approach aims to assign class penalties to shift the decision boundary in a way that reduces the bias induced by the data imbalance.
For this purpose, the most commonly adopted method is to re-weight samples inversely to the number of training samples in each class to assign more weights for the minority classes \cite{ref:huang_tang_cvpr16, ref:wang_hebert_nips17, ref:cui_belongie_cvpr19}.
These methods have focused on only global-level class distribution and assign the same fixed weight to all samples belonging to the same class.
However, not all samples in a dataset play an equal role in determining the model parameters~\cite{cook_influence}. 
That is, some samples have greater influences on forming a decision boundary. 
Hence, each sample needs to be re-weighted differently according to its impact on the model.

Recently, numerous studies have been conducted in which each sample is considered to design sample-wise loss functions~ \cite{ref:dong_zhu_iccv2017, ref:lin_focal_loss_iccv17, ref:malisiewicz_iccv2011}. 
Specifically, these methods down-weight well-classified samples and assign more weights to \textit{hard examples}, which yield high errors.
This re-weighting might lead to the complete training when the high capacity of DNNs is sufficient to finally memorize the whole training data \cite{ref:zhang_vinyals_iclr17, ref:arpit_memoriz_icml2017}.
This implies that DNN is overfitted to hard samples, which are located at the overlapping region between the majority and minority classes. 
In the imbalanced data, most hard samples are majority samples that enforce the decision boundary to be complex and shift to the minority region.  
 
To address the aforementioned problem, in this paper, we propose a loss-sensitive method to down-weight samples that cause overfitting of a DNN trained with highly imbalanced data. 
To this end, we derive a formula that  measures how much each sample influences the complex and biased decision boundary.
To derive the formula, we utilize the influence function~\cite{cook_influence}, which has been widely used in robust statistics.
Using the derived formula, we design a novel loss function, called influence-balanced (IB) loss, that adaptively assigns different weights to samples according to their influence on a decision boundary.
Specifically, we re-weight the loss proportionally to the inverse of the influence of each sample.
Our method is divided into two phases: standard training and fine-tuning for influence balancing.
During the fine-tuning phase, the proposed IB loss alleviates the influence of the samples that cause overfitting of the decision boundary.

Through extensive experiments on multiple benchmark data sets, we demonstrate the validity of our method, and show that the proposed method outperforms the state-of-the-art cost-sensitive re-weighting methods.
Furthermore, since our IB loss is not restricted to a specific task, model, or training method, it can be easily utilized in combination with other recent data-level algorithms and hybrid methods for class-imbalance problems. 

The main contributions of this paper are as follows: 
\vspace{-1mm}
\begin{itemize}
    \item We discover that the existing loss-based loss methods can lead a decision boundary of DNNs to eventually overfit to the majority classes.
    \vspace{-2mm}
    \item We design a novel influence-balanced loss function to re-weight samples more effectively in such a way that the overfitting of the decision boundary can be alleviated.
    \vspace{-2mm}
    \item We demonstrate that simply substituting our proposed loss for the standard cross-entropy loss significantly improves the generalization performance on highly imbalanced data. 
\end{itemize}

\section{Related Work}

\subsection{Class Imbalance Learning}
\vspace{-1mm}
To solve the imbalanced learning problem, numerous studies have been conducted.
The research can be divided into three approaches: data-level, cost-sensitive re-weighting, and meta-learning approaches.

\textbf{Data-level approach.}
The data-level approach aims to directly balance the training data distributions by re-sampling (e.g., under-sampling the majority classes or over-sampling the minority classes) \cite{ref:resample_chawla_smote02, ref:resample_hulse_icml07} or generating synthetic samples \cite{ref:sankha_gamo_iccv2019}. 
However, under-sampling can lose some valuable information, and it is not applicable when the data imbalance between classes is significant.
Although over-sampling or data generation could be effective, these methods are susceptible to overfitting to certain repetitive samples, and often require a longer training time.

\textbf{Re-weighting approach.}
Cost-sensitive re-weighting methods assign different weights to samples to adjust their importance.
Commonly used methods include re-weighting samples inversely proportional to the number of the class \cite{ref:huang_tang_cvpr16, ref:wang_hebert_nips17} or the square root of class frequency \cite{ref:Mahajan_Weiss_eccv2018}.
Instead of heuristically using the number of classes, Cui et al. \cite{ref:cui_belongie_cvpr19} proposed using the effective number of samples. 
While these methods can successfully assign more weights to the minority samples, they assign the same weights to all samples belonging to the same class, regardless of each importance. 

To assign different weights to each sample according to its importance on the model, numerous methods were proposed for re-weighting samples based on their difficulties or losses \cite{ref:lin_focal_loss_iccv17, ref:dong_zhu_iccv2017, ref:malisiewicz_iccv2011}.
That is, these methods down-weight well-classified samples and assign more weights to hard examples.
These re-weighting methods might cause DNNs to be overfitted to the hard examples, since the high capacity of DNNs is sufficient to memorize the training data in the end \cite{ref:arpit_memoriz_icml2017}.
In class imbalanced data, the hard examples are likely generated from the majority classes.
As such, the minority samples are assigned smaller weights.
Therefore, we need a more elaborate mean of re-weighting samples that can alleviate the overfitting to the majority samples. 
Meanwhile, Cao et al.~\cite{ref:cao_ldam_neurips2019} proposed label-distribution-aware margin loss to solve the overfitting to the minority classes by regularizing the margins.

\textbf{Meta-learning approach.}
Recently, the meta-learning-based approach~\cite{ ref:shu_metaweightnet_neurips2019, ref:liu_mesa_neurips20, ref:ren_bms_neurips2020} has emerged to enhance the performance of both approaches.
Shu et al.~\cite{ref:shu_metaweightnet_neurips2019} proposed a meta-learning process to learn a weighting function, while Liu et al.~\cite{ref:liu_mesa_neurips20} proposed a re-sampling method  by combining the advantage of ensemble learning and meta-learning. 
Furthermore, Ren et al.~\cite{ref:ren_bms_neurips2020} proposed the meta-sampler and a balanced softmax that accommodates the shift of the distributions between the training data and test data.
Although these methods can achieve satisfactory performance, these methods are somewhat difficult to implement in practice.
For example, meta-weight-net~\cite{ref:shu_metaweightnet_neurips2019} requires additional unbiased data for learning, and the meta-sampler in \cite{ref:ren_bms_neurips2020} is computationally expensive in practice.
On the other hand, our proposed loss is simple to implement because it does not require a hyperparameter, a specially designed architecture, or additional learning for data re-sampling.
Therefore, it is easy to use in collaboration with other methods.

\subsection{Influence function.}
\vspace{-1mm}
The influence function was proposed to find the influential instance of a sample to a model, which has been studied for decades in robust statistics \cite{ref:hampel1986robust, cook_influence}.
Recently, attempts have been made to use influence function in deep neural networks \cite{ref:FANN_2003, ref:Koh_Liang_2017}.
For example, Koh and Liang \cite{ref:Koh_Liang_2017} employed the influence function to understand DNNs.
While the influence function is primarily used as a diagnostic tool after the training of a model, 
our work first attempts to apply it to a learning scheme, in which we design the influence-balanced loss by utilizing the influence function during training.




\section{Method}
\vspace{-1mm}
To address the imbalanced data learning problem, our idea is to re-weight samples by their influences on a decision boundary to create a more generalized decision boundary.
First, we present the key idea of our proposed method in Section \ref{sec:idea}.
For the background, we briefly review the influence function in Section \ref{sec:influence} and then derive the IB loss in Sections \ref{sec:ibfactor}, \ref{sec:ib_loss}, and \ref{sec:class_reweight}.
Finally, the training scheme is presented in Section \ref{sec:training}.

\begin{figure}[t]
\begin{center}
\subfigure[Original decision boundary.]{\includegraphics[width=0.45\linewidth]{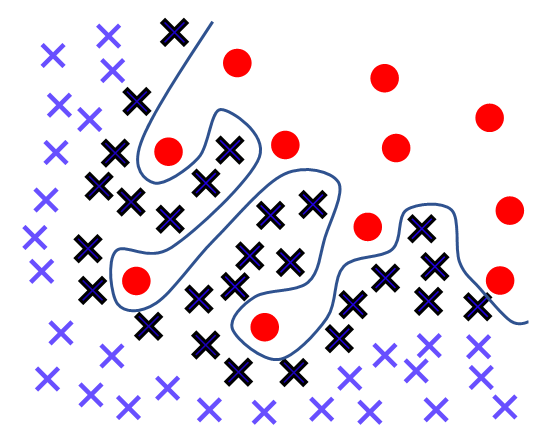}}
\hspace{0.2cm}
\subfigure[Proposed method.]{\includegraphics[width=0.45\linewidth]{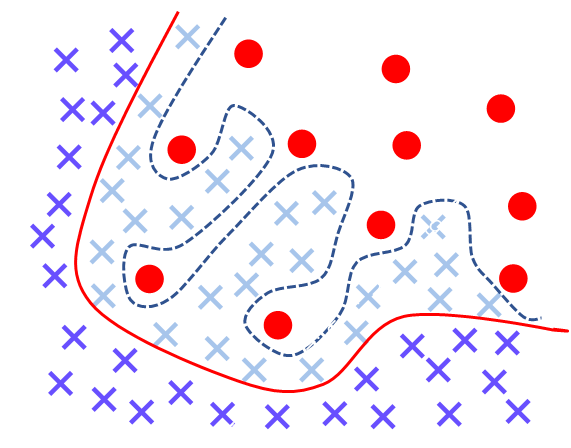}}
\end{center}
\caption{\textbf{Illustration of the key concept of our approach.} The red and blue marks belong to the minority and majority classes, respectively, in binary classification.
(a) The black border line represents an initial decision boundary formed on an imbalanced dataset. 
The black $\times$  samples have greater influence on the decision boundary than do the blue $\times$  samples, since the decision boundary would substantially change without the black $\times$  samples.
(b) 
Our proposed method aims to down-weight the samples (light blue $\times$ samples) that have a large influence on the overfitted decision boundary (dotted line) to create a smoother decision boundary (the red line). 
}
\label{fig:intuition}
\vspace{-5mm}
\end{figure}

\subsection{Key Idea of Proposed Method}\label{sec:idea}
\vspace{-1mm}
In this section, we explain how the re-weighting of samples according to their influence can help to form a  well-generalized decision boundary on class imbalance data.
It is well known that the high capacity of DNNs is sufficient to finally memorize the entire training data \cite{ref:zhang_vinyals_iclr17, ref:arpit_memoriz_icml2017}.
This implies that DNN can be overfitted to samples that are located at the overlapping region between the majority and minority classes, as illustrated in Figure \ref{fig:intuition} (a). 
In the imbalanced data, many majority samples invade among sparse minority samples and become dominant in the overlapping area, thereby enforcing the decision boundary to be complex and shift to the minority region.  

Furthermore, the black $\times$ samples in Figure \ref{fig:intuition} (a) have a stronger influence on forming the decision boundary, as they support the decision boundary, which substantially changes when the samples are removed. 
Thus, it can be said that the dominant samples with high influence are likely to create a complex and biased decision boundary. 
As illustrated in Figure \ref{fig:intuition} (b), by down-weighting the highly influential samples, the decision boundary can be smoothed via  fine-tuning.
To this end, we derive an influence-balanced (IB) loss by employing the influence function~\cite{cook_influence}, which measures the training sample's influence on the model.

\subsection{Influence Function}\label{sec:influence}
\vspace{-1mm}
The influence function \cite{cook_influence} allows us to estimate the change in the model parameters when a sample is removed, without actually removing the data and retraining the model.
Let $f(x, w)$ denote a model parameterized by $w$ with $n$ training data $(x_1, y_1), \cdots , (x_n, y_n)$, where $x_i$ is the $i$-th training sample, and $y_i$ is its label.
Given the empirical risk $R(w) = \frac{1}{n}\sum_{i=1}^n L(y_i, f(x_i, w))$, the optimal parameter after initial training is defined by ${w}^* \eqdef  \text{argmin}_{w} R(w)$.

During the fine-tuning phase, to address the imbalance issue, we re-weight loss proportionally to the inverse of the influence of a sample. 
The influence of a point $(x, y)$ can be approximated by the parameter change if the distribution of the training data at that point is slightly modified.
A new parameter when removing the training point $(x, y)$ is derived as ${w}_{x, \varepsilon} \eqdef  \text{argmin}_{w} R(w) + \varepsilon L(y, f(x, w)).$
Then, under the assumption that $\triangledown_{w}R(w) \approx 0$ for $w$ in the vicinity of $w^*$, we can utilize the influence function in \cite{ref:FANN_2003, ref:Koh_Liang_2017} to re-weight the sample-wise loss during the fine-tuning phase. The influence function is given by 
\vspace{-1mm}
\begin{equation}
    {\cal I}(x;w) = -H^{-1}\triangledown_{w}
    L(y, f(x, {w})), \label{eq:Iparams}
    \vspace{-1mm}
\end{equation}
where $H \eqdef \frac{1}{n}\sum_{i=1}^n \triangledown_{w}^2L(y_i, f(x_i, w))$ is the Hessian and is positive definite based by assumption that $L$ is strictly convex in a local convex basin around the optimal point ${w}^*$. 

\subsection{Influence-balanced weighting factor}\label{sec:ibfactor}
\vspace{-0.5mm}
From ${\cal I}(x;w)$, we derive the IB loss.
Since ${\cal I}(x;w)$ is a vector that requires heavy computation of the inverse Hessian, it is nearly impossible to directly use this.
Therefore, we solve this problem by modifying ${\cal I}(x;w)$ to a simple but effective influence-balanced weighting factor. 
First, since we need the relative influence of the training samples, not the absolute values, we can simply ignore the inverse Hessian in ${\cal I}(x;w)$.
This is because the inverse of hessian is commonly multiplied by all the training samples.
Then, we design the IB weighting factor as follows:
\vspace{-1mm}
\begin{equation}\label{eq:ibfactor}
    {\cal IB}(x;w) = ||\triangledown_{w} L(y, f(x, {w}))||_1
    \vspace{-1mm}
\end{equation}

Equation \ref{eq:ibfactor} turns out to be the magnitude of the gradient vector.
Anand et al. \cite{ref:anand_ranka_1993} revealed that the net error gradient vector is dominated by the major classes in the class imbalance problem.
Hence, re-weighting samples by the magnitude of the gradient vector can successfully down-weight samples from dominant classes.
In the Experiments section, we justify the choice of the L1 norm. 
In the following section, we demonstrate how the IB weighting factor can be used with the actual loss.

\subsection{Influence-Balanced Loss}\label{sec:ib_loss}
\vspace{-1mm}
When using the softmax cross-entropy loss, Equation (\ref{eq:ibfactor}) can be further simplified.
The cross-entropy loss is denoted by $L(y,f(x, {w})) = -\sum_k^K y_k\log f_k$, where $y_k$ is a ground truth, and $f_k$ is the $k$-th output of the model $f(x, {w})$, with $K$ total classes.
Since we are interested in the overfitting on the decision boundary of the model, we focus on the change in the last fully connected (FC) layer of a deep neural network.
Let ${h} = [h_1, \cdots, h_L]^T$ be a hidden feature vector,  an input to the FC layer, 
and $f(x, {w}) = [f_1, \cdots, f_K]^T$ be the output denoted by 
$f_k := \sigma({w}_k^T {h})$, where $\sigma$ is the softmax function. 
The weight matrix of the FC layer is denoted by $w = [{w}_1, \cdots , {w}_K]^T \in R^{K \times f}$. 

Then, the gradient of the loss w.r.t. $w_{kl}$ is computed as 

\begin{equation}
\begin{split}
    \frac{\partial }{\partial w_{kl}}L(y, f(x, {w}))
    = (f_k - y_k)h_l.
\end{split}
\end{equation}
The same results are obtained for the cross-entropy loss with a sigmoid function or a mean squared error (MSE) loss for regression. 
Then, IB weighting factor in (\ref{eq:ibfactor}) is given by 
\vspace{-1mm}
\begin{equation}
\vspace{-1mm}
\begin{split}
    {\cal IB}(x;w) &= \sum_k^K \sum_l^L |(f_k - y_k)h_l| \\
    &=\sum_k^K |(f_k - y_k)|\sum_l^L |h_l|\\
    &=||f(x,w)-y||_1 \cdot ||h||_1 ,
\end{split}
\end{equation}
of which inverse can be used for the re-weighting factor to down-weight an influential sample in fine-tuning to adjust the decision boundary that enhance the imbalanced data learning. Finally, the influence-balanced loss is given by 
\begin{equation}\label{eq:ibce}
    L_{IB}(y, f(x, {w}) = \frac{L(y,f(x, {w}))}{||f(x,w)-y||_1 \cdot ||h||_1}.
\end{equation}
The proposed influence-balanced term constrains the decision boundary to not overfit to influential majority samples (see Figure \ref{fig:intuition}(b)).
\setlength{\textfloatsep}{1pt}
\begin{algorithm}[h]
    \caption{Influence-Balanced Training}
    \label{algo:ib_training}
    \SetKwInOut{Input}{Input}
    \SetKwInOut{Output}{Output}
    \Input{training dataset $D = (X, Y)$.}
    \Output{influence-balanced model $f(x, {w})$.} \vspace*{1mm}
    {\bfseries Phase 1: Normal training}\\
    Initialize the model with random parameters $w$.\\
    \For{$t=1$ \KwTo $T_1$}{
    sample mini-batch $D_m$ from $D$ \\
    $L(w) \leftarrow \frac{1}{m}\sum_{(x, y)\in D_m} L(y, f(x, w))$\\
    update $w^t = w^{t-1} - \eta \triangledown L(w) $}
    \vspace*{1mm}
    {\bfseries Phase 2: Fine-tuning for influence balancing}\\
    \For{$t=T_1+1$ \KwTo $T$}{
    sample mini-batch $D_m$ from $D$ \\
    $L_{IB}(w) \leftarrow \frac{1}{m}\sum_{(x, y)\in D_m} \lambda_k \frac{L(y,f(x, {w}))}{||f(x,w)-y||_1 \cdot ||h||_1}$\\
    update $w^t = w^{t-1} - \eta \triangledown L(w) $}
\vspace{-1mm}
\end{algorithm}

\subsection{Influence-Balanced Class-wise Re-weighting}\label{sec:class_reweight}
\vspace{-0.5mm}
Moreover, we add a class-wise re-weighting term $\lambda_k$ to the IB-loss in (\ref{eq:ibce}) as
\vspace{-1mm}
\begin{equation}\label{eq:ibce_classwise}
    L_{IB}(w) = \frac{1}{m}\sum_{(x, y)\in D_m} \lambda_k  \frac{L(y,f(x, {w}))}{||f(x,w)-y||_1 \cdot ||h||_1},
\end{equation}
where 
$\lambda_k=\alpha n_k^{-1} / \sum_{k'=1}^K n_{k'}^{-1}$.
Here, $n_k$ is the number of samples in the $k$-th class in the training dataset, and normalization is performed to make $\lambda_k$ have a similar scale for every class. $\alpha$ is introduced as a hyper-parameter for an adjustment.

The class-wise re-weighting yields the following two effects.
First, $\lambda_k$ mitigates the bias of the decision boundary arising from the overall imbalanced distribution through the slow-down of the majority loss minimization. 
Second, $\lambda_k$ further controls the sample-wise re-weighting depending on the class to which a highly influential sample belongs.
That is, if the sample belongs to a majority class, $\lambda_k$ further down-weights the sample because the decision boundary is likely to be overfitted by the majority sample.
Meanwhile, if the sample belongs to a minority class, $\lambda_k$ becomes smaller than that of a majority sample and does not down-weight the loss much, because the large influence of the minority sample is natural due to the data scarcity. 

\vspace{-1mm}
\subsection{Influence-balanced Training Scheme}\label{sec:training}
\vspace{-1mm}
The influence-balanced training process comprises two phases: normal training and fine-tuning for balance.
We refer to $T_1$ as the transition time from normal training to fine-tuning. 
During the normal training phase, the network is trained following any training scheme for the first $T_1$ epochs.
Meanwhile, during the fine-tuning phase, the influence-balanced loss is applied to mitigate the overfitting of the decision boundary arising from the influential (noisy) majority samples. 
Since our IB loss during the fine-tuning phase alleviates the overfitting, it is advantageous to set $T_1$ as the epoch when the model has begun to converge to the local (global) minimum.
Generally, it is recommended to set $T_1$ as half of the total training scheme. 
We present the performance change according to the number of training epochs during normal training in the Experiments section. 
As evident, our training does not require an additional training scheme or a specifically designed architecture.
Thus, it can be utilized easily in any tasks suffering from  imbalanced data.
The pseudo-code of the training procedure is presented in Algorithm \ref{algo:ib_training}.

\section{Experiments}

\subsection{Experimental Settings} \label{sub:exp_setting}
\textbf{Datasets.} 
We verified the effectiveness of our method on three commonly used benchmark datasets: CIFAR-10, CIFAR-100 \cite{ref:data_cifar}, Tiny ImageNet~\cite{data:tinyimagenet}, and iNaturalist 2018 \cite{ref:data_iNat}.
The CIFAR-10 and CIFAR-100 datasets consist of 50,000 training images and 10,000 test images with 10 and 100 classes, respectively. 
Meanwhile, Tiny ImageNet contains 200 classes for training, in which each class has 500 images. 
Its test set contains 10,000 images.
Since CIFAR and Tiny ImageNet are evenly distributed, we have made these datasets imbalanced according to \cite{ref:cui_belongie_cvpr19, ref:buda_mazurowski_2018}, respectively.
Primarily, we investigate two common types of imbalance: (i) long-tailed imbalance \cite{ref:cui_belongie_cvpr19} and (ii) step imbalance \cite{ref:buda_mazurowski_2018}.
In long-tailed imbalance, the number of training samples for each class decreases exponentially from the largest majority class to the smallest minority class. 
To construct long-tailed imbalanced datasets, the number of selected samples in the $k$-th class was set to $n_k \mu ^ k (\mu \in (0, 1) )$, where $n_k$ is the original number of the $k$-th class.
Meanwhile, in step imbalance, the classes are divided into two groups: the majority class group and minority class group. 
Every class within a group contains the same number of samples, and the class in the majority class group has many more samples than that in the minority class group.  
For evaluation, we used the original test set. 
The imbalance ratio $\rho$ is defined by $\rho = \frac{\max_k \{n_k\}}{\min_k \{n_k\}}$.
Thus, the imbalance ratio represents the degree of imbalance in the dataset. 
We evaluated the performance of our method under various imbalance ratios from 10 to 200.

The iNaturalist 2018 dataset is a large-scale real-world dataset containing 437,513 training images and 24,426 test images with 8,142 classes.
iNaturalist 2018 exhibits long-tailed imbalance, whose imbalance ratio is 500.
We used the official training and test splits in our experiments. \vspace{2mm}

\textbf{Baselines.}
We compared our algorithm with the following cost-sensitive loss methods:
(1) Our baseline model, which is trained on the standard cross-entropy loss. 
Comparing our model with this baseline enables us to clearly understand how much our training scheme has improved the performance;
(2) focal loss \cite{ref:lin_focal_loss_iccv17}, which increases the relative loss for hard samples and down-weights well-classified samples;
(3) CB loss \cite{ref:cui_belongie_cvpr19}, which re-weights the loss inversely proportional to the effective number of samples;
(4) LDAM loss \cite{ref:cao_ldam_neurips2019}, which regularizes the minority classes to have larger margins.  

Since our IB loss can be easily combined with other methods, we employee two further variants.
First, IB + CB uses the effective number in CB loss, instead of using $\lambda_k$ in IB.
Second, IB + focal uses focal loss during the fine-tuning phase, instead of using the cross-entropy loss.
We demonstrate that combination with other methods can further improve the performance.

\textbf{Implementation Details.} \label{sub:impl_detail}
We used PyTorch \cite{ref:Pytorch} to implement and train all the models in the paper, and we used ResNet architecture \cite{ref:resnet_cvpr2016} for all datasets. 
For CIFAR datasets, we used randomly initialized ResNet-32.
The networks were trained for 200 epochs with stochastic gradient descent (SGD) (momentum = 0.9).
Following the training strategy in \cite{ref:cui_belongie_cvpr19, ref:cao_ldam_neurips2019}, the initial learning rate was set to 0.1 and then decayed by 0.01 at 160 epochs and again at 180 epochs. 
Furthermore, we used a linear warm-up of the learning rate \cite{ref:goyal_he_corr2017} in the first five epochs.
Since our method uses a two-phase training schedule, we trained for the first 100 epochs with the standard cross-entropy loss, then fine-tuned the networks using the IB loss for the next 100 epochs.
We trained the models for CIFAR on a single NVIDIA GTX 1080Ti with a batch size of 128.
For Tiny ImageNet, we employed ResNet-18 and used the stochastic gradient descent with a momentum of 0.9, and weight decay of $2\mathrm{e}{-4}$ for training. 
The networks were initially trained for 50 epochs, and then fine-tuned for the subsequent 50 epochs with IB loss.
The learning rate at the start was set to 0.1 and was dropped by a factor of 0.1 after 50 and 90 epochs.
For iNaturalist 2018, we trained ResNet-50 with four GTX 1080Ti GPUs.
The networks were initially trained for 50 epochs and then fine-tuned for the subsequent 150 epochs with IB loss.
The learning rate at the start was set to 0.01 and was decreased by a factor of 0.1 after 30 and 180 epochs.

As a simple but important implementation trick, we added $\epsilon = 0.001$ to ${\cal IB}(x;w)$ to prevent numerical instability in inversion when the influence approaches zero.
We discuss the influence of the hyperparameter ($\epsilon$) in the following section.

\subsection{Analysis}
To validate the proposed method, we conducted extensive experiments.

\begin{figure}[t]
\begin{center}
\includegraphics[width=0.9\linewidth]{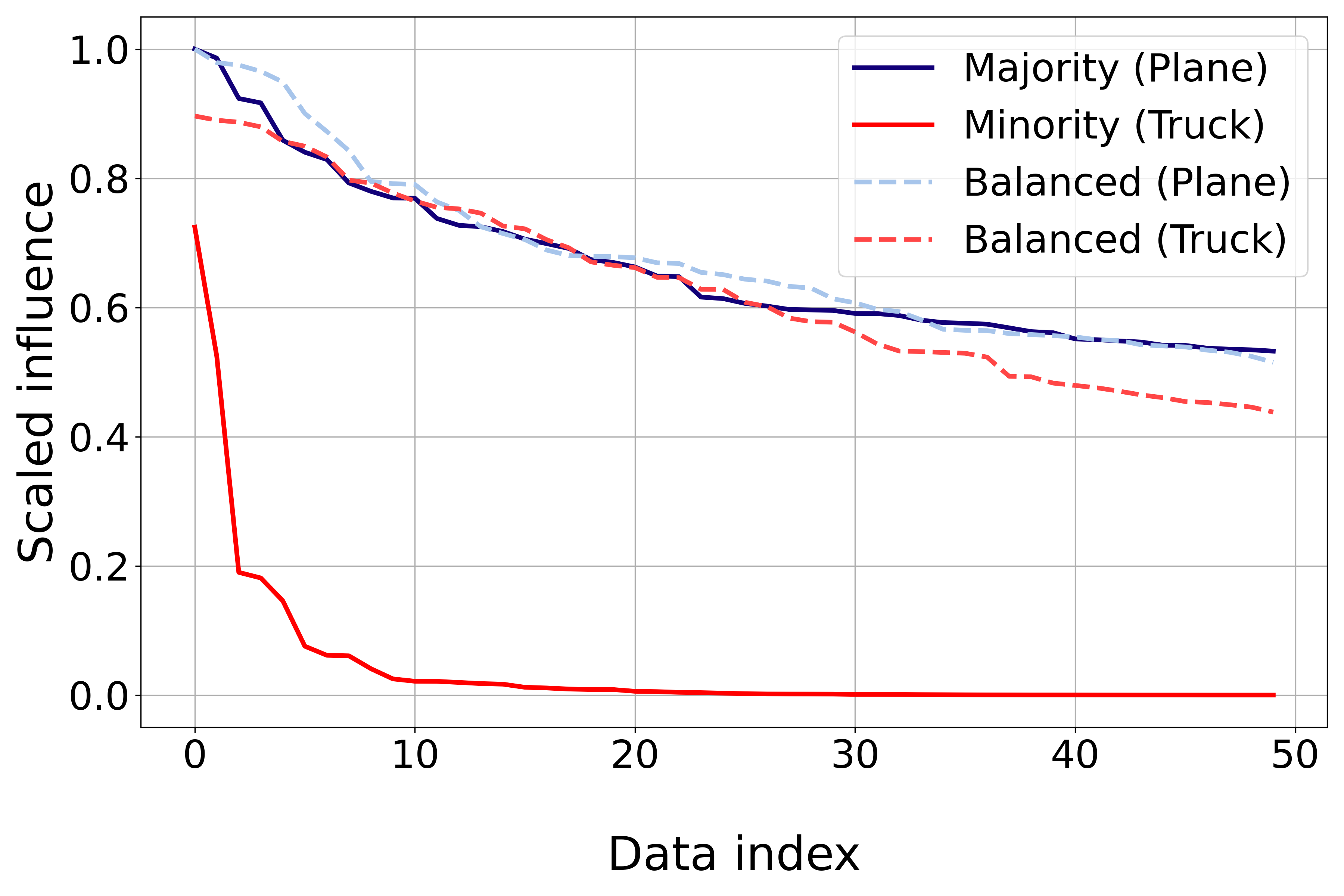}
\caption{\textbf{Comparison of Influences between balanced and imbalanced dataset.}
We plotted the influences of samples on ResNet-32 trained on the original CIFAR-10 and the imbalanced version of CIFAR-10. The solid and dashed lines represent the influences of the imbalanced data and balanced data, respectively.
While there is little difference in the balanced dataset, it can be seen that the influence of the dominant class is much greater than that of the minor class in the imbalance dataset.
} 
\label{fig:bal_imbal}
\end{center}
\end{figure}

\textbf{Is influence meaningful for re-weighting?}
First, to confirm whether influence can act as a meaningful clue of re-weighting for class imbalance learning, we compared the influences between a balanced dataset and an imbalanced dataset.
For an imbalanced CIFAR-10, we used the long-tailed version of CIFAR-10 with the imbalance ratio $\rho = 100$, in which
the largest class, `plane' (i.e., class index 0), contains 5,000 samples, while the smallest class, `truck' (i.e., class index 9), contains only 50 samples.
We trained ResNet-32 with a standard cross-entropy loss for 200 epochs, as described in Implementation Details, on both the balanced (original) and imbalanced CIFAR-10.
We plotted the influences of both classes in Figure \ref{fig:bal_imbal}.
We scaled the influences to between 0 and 1 for each dataset. 
Since the minority class contains only 50 samples, we selected the highest 50 samples for comparison. 
As illustrated in Figure \ref{fig:bal_imbal}, there was little difference in the distributions of the influences between the classes in the balanced dataset.
However, in the imbalanced dataset, the minority samples had significantly less influence on the model than did the majority samples.
This result corroborates that majority samples greatly contribute to forming a decision boundary, and re-weighting their influences can improve the generalization of the model.

\textbf{Magnitude of Influence.}
In Section \ref{sec:ibfactor}, we used $L_1$ norm to compute the magnitude of the influences.
We investigated performance variations depending on three vector norms to compute the magnitude of the gradient vector $\triangledown_{w} L(y, f(x, {w}))$: $L_1$, $L_2$, $L_\infty$.
As indicated in Table \ref{ref:table_norms}, $L_1$ norm, which provides a distinctive change of influence around the equilibrium point, exhibits the best classification accuracy on CIFAR-10 with multiple imbalance ratios.

\begin{table}[h]
\caption{Comparison of norms. Using $L_1$ norm yields the best performance.}
\small
\centering
\begin{tabular}{ccccc}
\toprule
\multirow{2}{*}{} & \multicolumn{2}{c}{\textbf{CIFAR-10}} & \multicolumn{2}{c}{\textbf{CIFAR-100}} \\ \cmidrule(lr){2-3}\cmidrule(lr){4-5}
Imbalance ($\rho$)            & $100$          & $20$         & $100$          & $20$     \\ \midrule\midrule
$L_1$             & \textbf{78.41} & \textbf{85.80}& \textbf{40.85} & \textbf{52.85}         \\ 
$L_2$             & 75.67          & 84.35        & 36.41          & 50.95         \\ 
$L_\infty$        & 77.23          & 84.30         & 37.48          & 50.99         \\ 
\bottomrule
\end{tabular}
\label{ref:table_norms}
\vspace{-2mm}
\end{table}

\begin{figure}[h]
\vspace{-0.2cm}
\begin{center}
\includegraphics[width=0.95\linewidth]{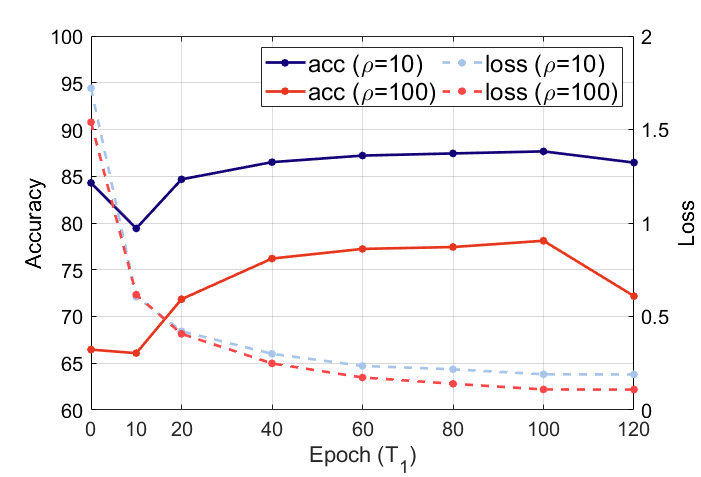}
\caption{\textbf{Influence-balanced training scheme.}
We varied the \textit{training epochs for the normal training}, $T_1$, to determine the best transition time from the normal training to the influence-balance fine-tuning. 
We achieved the best performance 
when setting the transition time to the point when the training loss converges.
}
\label{fig:training_epochs}
\end{center}
\end{figure}

\textbf{Timing for starting fine-tuning for balancing.}
Our training scheme is divided into two phases: normal training and fine-tuning for balancing. 
This must determine the transition time between normal training and fine-tuning for balancing. 
Hence, we investigated the results on how much the transition time affects the performance and determined the best transition time.
For this, we experimented on the long-tailed version of  CIFAR-10 with imbalance ratios of $\rho = 10$ and $100$.
In Figure \ref{fig:training_epochs}, the $X$-axis represents the number of training epochs $T_1$ for the normal training phase.
We varied the transition time, $T_1$, from 0 to 120 while the total number of training epochs was fixed at 200. 
The solid line represents the classification accuracy earned by the models for each training schedule.
To analyze the relationship between the convergence of the normal training phase and the transition timing, we plotted the standard cross-entropy loss without adopting the IB loss for the whole training epochs (dashed lines).

From Figure \ref{fig:training_epochs}, it can be observed that the proposed method demonstrates robust performance regardless of the choice of transition time $T_1$.
Yet, the transition to fine-tuning after the 100th epoch yields the best performance when the training loss has converged.
Since the influence function is derived from the loss minimization context~\cite{ref:Koh_Liang_2017}, it is reasonable to begin the fine-tuning phase after the learning converges. 


\textbf{Effects of $\epsilon$.}
As mentioned in Implementation Details, for all datasets, we added the hyperparameter ($\epsilon = 0.001$) to ${\cal IB}(x;w)$ to prevent numerical instability.
To analyze the effects of the hyperparameter, we conducted experiments with the following denominators for the IB loss~(\ref{eq:ibce}):
(a) ${\cal IB}(x;w) + 1\mathrm{e}{-8}$, (b) ${\cal IB}(x;w) + 1\mathrm{e}{-3}$, (c) ${\cal IB}(x;w) + 1\mathrm{e}{-2}$, and (d) $1\mathrm{e}{-3}$. 
We iterated experiments three times with different random seeds on the long-tailed CIFAR-10 ($\rho = 100$).
As presented in Table~\ref{table:epsilon}, setting $\epsilon$ to $1\mathrm{e}{-3}$ yields the best performance.
Thus, we set  $\epsilon$ as $1\mathrm{e}{-3}$ in all the experiments.
However, when we did not use the IB weighting factor, the accuracy greatly decreased.

\begin{table}[h]
\caption{Effects of $\epsilon$.}
\centering
\resizebox{1\linewidth}{!}{
\begin{tabular}{lllll}
\toprule
Epsiilon & (a) IB+1e-8 & (b) IB+1e-3 & (c) IB+1e-2    & (d) 1e-3 \\ \midrule
Accuracy & $76.03 \pm 0.97$ & $78.17 \pm 0.57$ & $77.55 \pm 0.55$ &  $64.91 \pm 1.40$     \\ \bottomrule
\end{tabular}
\label{table:epsilon}
}\vspace{-0.4cm}

\end{table}

\begin{table*}[t]
\caption{
Class-wise classification accuracy (\%) of ResNet-32 on imbalanced CIFAR-10 dataset.
The number of test samples for each class is the same as 1000.
The best results are marked in bold. 
}
\vspace{0.2cm}
\label{result-class-wise}
\centering
\small
{
\begin{tabular}{
l
ccccc
ccccc
} 
    \toprule
    &  \multicolumn{10}{c}{\textbf{Imbalanced CIFAR-10}}
    \\
    \cmidrule(){2-11}
    Class 
    & plane & car & bird & cat & deer 
    & dog & frog & horse & ship & truck 
    \\
    \midrule\midrule
    
    \textit{\textbf{Long-Tailed}} ($\rho = 50$)
    &  &  &  &  & 
    &  &  &  &  & 
    \\
    \#Training samples
    & 5000 & 3237 & 2096 & 1357 & 878 
    & 568 & 368 & 238 & 154 & 100 
    \\
    \midrule
    Baseline (CE)
        & \textbf{97.4}
        & \textbf{98.0}
        & \textbf{84.0}
        & \textbf{80.3}
        & 78.8
        & 68.4
        & 76.1
        & 64.5
        & 57.0
        & 52.0
        \\
        
     Focal \cite{ref:lin_focal_loss_iccv17}
        & 91.6
        & 95.1
        & 73.1
        & 59.2
        & 67.8
        & 67.2
        & 84.2
        & 77.3
        & \textbf{83.9}
        & 61.8
        \\
    
    CB ~\cite{ref:cui_belongie_cvpr19}
        & 92.9
        & 96.3
        & 79.2
        & 75.1
        & 82.4
        & 69.9
        & 75.0
        & 69.1
        & 73.6
        & 66.8
        \\
        
    LDAM~\cite{ref:cao_ldam_neurips2019}
        & 96.9
        & 98.5
        & 82.9
        & 74.7
        & 82.8
        & 69.0
        & 78.5
        & 69.9
        & 65.3
        & 66.0
        \\
        
    LDAM-DRW~\cite{ref:cao_ldam_neurips2019}
        & 94.8
        & 97.8
        & 82.6
        & 72.3
        & 85.3
        & 73.0
        & 82.0
        & 76.7
        & 75.8
        & 72.4
        \\
    
    \midrule
    
    IB 
        & 92.2
        & 96.2
        & 81.3
        & 66.6
        & \textbf{85.7}
        & \textbf{76.4}
        & 81.7
        & 75.9
        & 79.9
        & \textbf{81.1}
        \\
        
    IB + CB
        & 93.8
        & 97.2
        & 78.1
        & 64.8
        & 84.8
        & 74.2
        & \textbf{86.4}
        & \textbf{79.7}
        & 79.5
        & 76.9
        \\
    
    IB + Focal 
        & 90.9
        & 96.1
        & 81.7
        & 69.0
        & 82.0
        & 75.7
        & 85.2
        & 77.5
        & 80.2
        & 76.8
        \\
    
    \midrule\midrule
    
    \textit{\textbf{Step-Imbalance}} ($\rho = 50$)
    &  &  &  &  & 
    &  &  &  &  & 
    \\
    
    \#Training samples
    & 5000 & 5000 & 5000 & 5000 & 5000 
    & 100 & 100 & 100 & 100 & 100 
    \\
    \midrule
    
    Baseline (CE)
        & 95.9
        & \textbf{99.2}
        & \textbf{91.5}
        & \textbf{91.9}
        & 95.5
        & 24.8
        & 40.2
        & 46.7
        & 52.7
        & 55.1
        \\
        
    Focal \cite{ref:lin_focal_loss_iccv17}
        & 96.3
        & 93.9
        & 91.2
        & 90.5
        & \textbf{95.7}
        & 20.0
        & 46.7
        & 48.8
        & 56.1
        & 57.6
        \\
    
    CB ~\cite{ref:cui_belongie_cvpr19}
        & 87.4
        & 96.3
        & 76.8
        & 77.0
        & 85.7
        & 34.6
        & 61.5
        & 56.5
        & 68.7
        & 63.8
        \\
        
    LDAM~\cite{ref:cao_ldam_neurips2019}
        & \textbf{96.4}
        & 98.5
        & 91.1
        & 90.2
        & 94.6
        & 28.3
        & 50.3
        & 57.0
        & 56.2
        & 64.4
        \\
        
    LDAM-DRW~\cite{ref:cao_ldam_neurips2019}
        & 94.5
        & 97.2
        & 88.0
        & 84.5
        & 94.3
        & 50.4
        & 69.9
        & 71.4
        & 74.6
        & 76.0
        \\
    
    \midrule
    
    IB 
        & 94.0
        & 97.7
        & 86.7
        & 83.2
        & 93.8
        & 56.9
        & 71.0
        & \textbf{75.1}
        & 76.5
        & 81.7
        \\
        
    IB + CB
        & 91.8
        & 95.7
        & 86.6
        & 79.4
        & 93.6
        & 62.8
        & 77.2
        & 72.3
        & 74.2
        & \textbf{87.3}
        \\
    
    IB + Focal 
        & 91.2
        & 96.4
        & 83.3
        & 77.1
        & 92.0
        & \textbf{64.8}
        & \textbf{78.0}
        & 74.4
        & \textbf{83.5}
        & 83.1
        \\

    \bottomrule
\end{tabular}
\vspace{-2mm}
}
\end{table*}

\subsection{Comparison of Class-Wise Accuracy.}
\vspace{-1mm}
In this section, to validate that the performance improvement has actually resulted from the minority classes, not from the majority classes, we report the class-wise accuracy on both the long-tailed and the step-imbalanced CIFAR-10.
We compare the proposed method with the state-of-the-art cost-sensitive loss methods. 
Since previous studies do not report the class-wise accuracy on the imbalanced CIFAR-10, we implemented the baseline methods~ \cite{ref:lin_focal_loss_iccv17, ref:cui_belongie_cvpr19, ref:cao_ldam_neurips2019}. 
For the implementation of LDAM~\cite{ref:cao_ldam_neurips2019}, we used their official implementation code to reproduce the results. 

The overall results are reported in Table \ref{result-class-wise}.
As presented in Table~\ref{result-class-wise}, existing methods exhibit severe performance degradation in the minority classes.
That is, the reported improvements from the existing methods were attributed to the majority classes, not the minority classes.
In contrast, the proposed IB loss exhibited a significant improvement in all the minority classes. 

It is noteworthy that the performance improvement was not significant, especially on the step-imbalanced CIFAR-10 with the focal loss~\cite{ref:lin_focal_loss_iccv17} method.
We argue that this demonstrates that most hard examples are majority samples in highly imbalanced data and that those samples enforce the decision boundary to be overfitted.
In contrast, our proposed influence-balanced re-weighing can mitigate the influences of the majority samples that cause overfitting.
As a result, it can achieve robust and superior performance for the minority classes with a very small number of samples.

Although using the influence-balanced loss alone can achieve significant enhancement for the classification of the minority classes, it is beneficial to combine it with other methods. 
For example, the results indicate that applying the influence-balanced loss with the focal loss can encourage the network to learn `good' hard samples, while down-weighting the influential ones that induce overfitting. 

\begin{table*}[t]
\caption{
Classification accuracy (\%) of ResNet-32 on imbalanced CIFAR-10 and CIFAR-100 datasets.
``$\dagger$" indicates that the results are copied from the original paper, 
and ``$\ddagger$" means that the results are from the experiments in CB~\cite{ref:cui_belongie_cvpr19}.
The best results are marked in bold.
}
\vspace{0.2cm}
\label{result-table}
\centering
\small
{
\begin{tabular}{
l
ccccc
ccccc
} 
    \toprule
    &  \multicolumn{5}{c}{\textbf{Imbalanced CIFAR-10}}  
    &  \multicolumn{5}{c}{\textbf{Imbalanced CIFAR-100}} 
    \\
    \cmidrule(lr){2-6}\cmidrule(lr){7-11}
    Imbalance ($\rho$)
    & 200 & 100 & 50 & 20 & 10 
    & 200 & 100 & 50 & 20 & 10 
    \\
    \midrule\midrule
    \textit{\textbf{Long-Tailed}}
    &  &  &  &  & 
    &  &  &  &  & 
    \\
    Baseline (CE)
        & 66.28	
        & 70.87	
        & 78.22	
        & 82.43
        & 86.49
        & 33.54	
        & 38.05	
        & 43.71	
        & 51.21	
        & 56.96
        \\
        
    $^{\ddagger}$Focal~\cite{ref:lin_focal_loss_iccv17}
        & 65.29	
        & 70.38 	
        & 76.71 
        & 82.76 
        & 86.66 
        & 35.62
        & 38.41 
        & 44.32 
        & 51.95
        & 55.78
        \\
    
    $^{\dagger}$CB ~\cite{ref:cui_belongie_cvpr19}
        & 68.89	
        & 74.57	
        & 79.27	
        & 84.36	
        & 87.49
        & 36.23	
        & 39.60	
        & 45.32	
        & 52.59	
        & 57.99
        \\
        
    $^{\dagger}$LDAM~\cite{ref:cao_ldam_neurips2019}
        & -
        & 73.35
        & -
        & -
        & 86.96
        & -
        & 39.6
        & -
        & -
        & 56.91
        \\
    
    $^{\dagger}$LDAM-DRW~\cite{ref:cao_ldam_neurips2019}
        & -	
        & 77.03	
        & -	
        & -	
        & 88.16
        & -	
        & 42.04	
        & -	
        & -	
        & 57.99
        \\
    
    \midrule
    IB 
        & 73.96	
        & 78.26	
        & \textbf{81.70}
        & \textbf{85.8}	
        & \textbf{88.25}
        & 37.31	
        & \textbf{42.14}	
        & 46.22	
        & 52.63	
        & 57.13
        \\
    
    IB + CB 
        & 73.69	
        & 78.04	
        & 81.54	
        & 85.42
        & 88.09
        & 37.06	
        & 41.31	
        & 46.16	
        & 52.74	
        & 56.78
        \\
        
    IB + Focal 
        & \textbf{75.05}	
        & \textbf{79.76	}
        & 81.51	
        & 85.31	
        & 88.04
        & \textbf{38.23}	
        & 42.06	
        & \textbf{47.49}	
        & \textbf{53.28}	
        & \textbf{58.20}
        \\
    
    \midrule\midrule
    
    \textit{\textbf{Step-Imbalance}}
    &  &  &  &  &  
    &  &  &  &  &  
    \\
    
    Baseline (CE)
        & 56.97	
        & 64.81	
        & 69.35
        & 79.71	
        & 84.16
        & 38.29	
        & 39.27	
        & 41.65	
        & 48.55	
        & 54.13
        \\
    
    $^{\dagger}$LDAM~\cite{ref:cao_ldam_neurips2019}
        & -
        & 66.58
        & -
        & -
        & 85.00
        & -
        & 39.58
        & -
        & -
        & 56.27
        \\
        
    $^{\dagger}$LDAM-DRW~\cite{ref:cao_ldam_neurips2019}
        & -	
        & 76.92	
        & -	
        & -	
        & 87.81
        & -	
        & 45.36	
        & -	
        & -	
        & 59.46
        \\
        
    \midrule
    IB 
        & 72.15	
        & 76.53	
        & 81.66	
        & 85.41	
        & 87.72
        & 39.66	
        & \textbf{45.39}	
        & \textbf{48.93}	
        & 53.57	
        & 57.96
        \\
    
    IB + CB 
        & 69.96	
        & 75.97	
        & 82.09	
        & 85.27	
        & \textbf{88.01}
        & 39.69	
        & 45.27	
        & 48.80
        & 53.42	
        & 57.86
        \\
        
    IB + Focal 
        & \textbf{74.12}	
        & \textbf{77.97}	
        & \textbf{82.38}	
        & \textbf{85.68}	
        & 87.90
        & \textbf{40.39}	
        & 44.96	
        & 48.92	
        & \textbf{54.53}	
        & \textbf{59.54}
        \\
    
    \bottomrule
\end{tabular}
}
\vspace{0.2cm}
\end{table*}

\subsection{Comparison with State-of-the-Art}
\vspace{-1mm}

\textbf{Experimental results on CIFAR.}
The overall classification accuracy is provided in Table \ref{result-table}.
The model performance is reported on the unbiased test set as the same as the other methods.
The results indicate that adopting the proposed influence-balanced loss significantly improves the generalization performance and outperforms the recent cost-sensitive loss methods. 
On multiple benchmark datasets, using IB loss alone could achieve the best performance.
This suggests that it is effective for the robustness of the model to balance the influence of samples responsible for overfitting the decision boundary.
When combined with other methods~\cite{ref:cui_belongie_cvpr19, ref:lin_focal_loss_iccv17}, we could further improve the accuracy on multiple datasets.
This indicates that our proposed method of down-weighting influential samples that induce overfitting can benefit other methods as well. \vspace{1mm}

\textbf{Experimental results on Tiny ImageNet.}
We evaluated our method on Tiny ImageNet.
While we performed the experiments for the other baselines, the results of LDAM were copied from their original paper.
As presented in Table \ref{TinyImageNet}, IB loss outperforms other baselines on Tiny ImageNet as well.

\textbf{Experimental results on iNaturalist 2018.}
We evaluated our method on the large-scale real-world image data, iNaturalist 2018.
We compared our method with the state-of-the-art loss-based methods. 
Table~\ref{table:result_inat} reveals that simply balancing the influence of loss could achieve considerable improvement.

\section{Conclusion}

In this paper, we propose a novel influence-balanced loss to solve the overfitting of the majority classes in a class imbalance problem.
A model trained on imbalanced class data is susceptible to overfitting due to the high capacity of DNN and the scarcity of samples in certain classes. 
Therefore, as learning progresses, existing methods are likely to produce undesirable results, such as assigning higher weights to samples from majority classes. 
Unlike the existing methods, IB loss can robustly assign weights because it directly focuses on a sample's influence on the model.
We conducted experiments to demonstrate that our method can improve generalization performance under a class imbalance setting. 
In addition, our method is easy to be implemented and integrated into existing methods.
In the future, we plan to extend our method by incorporating data-level methods or other recent meta-learning methods. 

\vspace{-2mm}
\begin{table}[h]
\label{TinyImageNet}
\caption{Class. accuracy (\%) of ResNet-18 on Tiny ImageNet.}
\centering
\small
\begin{tabular}{l
cccc
}
\toprule
&  \multicolumn{2}{c}{\textit{\textbf{Long-Tailed}}}  
&  \multicolumn{2}{c}{\textit{\textbf{Step-Imbalance}}} 
\\
    \cmidrule(lr){2-3}\cmidrule(lr){4-5}
    Imbalance ($\rho$)
    & 100 & 10 
    & 100 & 10 
    \\
 \midrule\midrule
    Baseline (CE)
        & 38.52        & 36.62        & 36.74        & 51.11 \\
Focal~\cite{ref:lin_focal_loss_iccv17}                          & 38.95           & 54.02          & 38.24            & 41.77           \\
CB~\cite{ref:cui_belongie_cvpr19}                              & 41.37           & 54.82          & 37.35            & 54.3           \\
LDAM*~\cite{ref:cao_ldam_neurips2019}                           & 37.47           & 52.78          & 39.37            & 52.57           \\
IB                              & \textbf{42.65}           & \textbf{57.22 }         & \textbf{41.13}            & \textbf{54.83}            \\
\bottomrule
\end{tabular}
\label{table:result_tiny}
\end{table}

\vspace{-1cm}
\begin{table}[h]
\caption{Class. accuracy (\%) of ResNet-50 on iNaturalist 2018.}
\centering
\small
\begin{tabular}{l
cc
}
\toprule
\multirow{2}{*}{} & \multicolumn{2}{c}{\textbf{iNaturalist 2018}}                  \\ \cmidrule(lr){2-3}
                  Method & top1 & top5 \\ \midrule\midrule
Baseline (CE)     & 57.30                      & 79.48                     \\ 
Focal~\cite{ref:lin_focal_loss_iccv17}     & 58.03                      & 78.65                     \\ 
CB~\cite{ref:cui_belongie_cvpr19}                & 61.12                     & 81.03                     \\ 
LDAM~\cite{ref:cao_ldam_neurips2019}              & 64.58                     & 83.52                     \\ \midrule
IB                & \textbf{65.39}            & \textbf{84.98}             \\ \bottomrule
\end{tabular}
\label{table:result_inat}
\vspace{1cm}
\end{table}

\vspace{1cm}
\section*{Acknowledgement}
\vspace{-1mm}
This work was supported by Institute of Information \& Communications Technology Planning \& Evaluation(IITP) grants funded by the Korea government(MSIT) (No.B0101-15-0266, Development of High Performance Visual BigData Discovery Platform for Large-Scale Realtime Data Analysis) and (2017-0-00306, Multimodal sensor-based intelligent systems
for outdoor surveillance robots).

{\small
\bibliographystyle{ieee_fullname}
\bibliography{egbib}
}

\end{document}